\documentclass[letterpaper, 10 pt, conference]{ieeeconf}  

\IEEEoverridecommandlockouts                              

\usepackage{graphicx}
\usepackage{balance}
\usepackage{comment}
\usepackage{cite}
\usepackage{amssymb}
\usepackage[tight,footnotesize]{subfigure}
\usepackage[active]{srcltx}
\usepackage{amsmath}
\usepackage{amsfonts}

\newcommand{\norm}[1]{\left\lVert#1\right\rVert}
\graphicspath{{./figs/}}
\usepackage{color,soul}
\usepackage{eurosym}
\usepackage{algorithm}
\usepackage{algorithmic}
\usepackage{multicol}
\usepackage{xcolor}
\UseRawInputEncoding
\usepackage{esvect}
\usepackage{bm}
\usepackage{dsfont}

\title{\LARGE \bf
Cooperative Simultaneous Tracking and Jamming\\ for Disabling a Rogue Drone
}

\author{Savvas~Papaioannou,~Panayiotis~Kolios,~Christos~G.~Panayiotou and\\ ~Marios~M.~Polycarpou% <-this % stops a space
\thanks{The authors are with the KIOS Research and Innovation Centre of Excellence (KIOS CoE) and the Department of Electrical and Computer Engineering, University of Cyprus, Nicosia, 1678, Cyprus. E-mail:{\tt\small \{papaioannou.savvas, pkolios, christosp, mpolycar\}@ucy.ac.cy}%
}}

\begin{document}

\maketitle

\begin{abstract}
This work investigates the problem of simultaneous tracking and jamming of a rogue drone in 3D space with a team of cooperative unmanned aerial vehicles (UAVs). We propose a decentralized estimation, decision and control framework in which a team of UAVs cooperate in order to a) optimally choose their mobility control actions that result in accurate target tracking and b) select the desired transmit power levels which cause uninterrupted radio jamming and thus ultimately disrupt the operation of the rogue drone. 
The proposed decision and control framework allows the UAVs to reconfigure themselves in 3D space such that the \textit{cooperative simultaneous tracking and jamming} (CSTJ) objective is achieved; while at the same time ensures that the unwanted inter-UAV jamming interference caused during CSTJ is kept below a specified critical threshold. Finally, we formulate this problem under challenging conditions i.e., uncertain dynamics, noisy measurements and false alarms. Extensive simulation experiments illustrate the performance of the proposed approach. 
\end{abstract}

%We assume that each of the UAVs is equipped with a 3D range-finding sensor and we use stochastic filtering to recursively estimate the rogue drone's state over time given uncertain dynamics, noisy measurements and false alarms. Finally, we show how this hard combinatorial problem can be approximated by simpler one and solved with binary programming. 

\section{Introduction} \label{sec:intro}

In the recent years the demand for consumer drones (i.e., UAVs) has been skyrocketed \cite{FAA2019}. This new gadget has become extremely appealing to the consumers and has nowadays become ubiquitous. Drones however, like every new emerging technology, can potentially introduce new threads and risks for the public safety. Indeed, consumer drones have created a big risk for public safety especially around airports and restricted airspaces. In fact, numerous times airports have been shut down \cite{Schneider2019} because of rogue drones, causing long delays to the flights schedule and inconvenience to the passengers.

Unfortunately, no adequate solution exist for this problem as of today. Various approaches developed in academic and industrial labs have been focused on a) drone detection techniques \cite{Guvenc2018} based on RF signal sniffing, computer vision and sensor fusion, and b) interception techniques \cite{Loeb2017} such as net-casting, RF denial systems, high-power laser and even trained eagles \cite{BBC2016}. According to \cite{Wesson2013}, solutions for safeguarding drones are still in their infancy and considerable more work is needed in order for this technology to reach the required level of maturity. 

This paper deals with the deployment of a team of UAVs (i.e., agents) in 3D space with the ultimate purpose of intercepting and downing a single rogue drone (i.e., target). In particular, in this work we propose a multi-UAV \textit{cooperative simultaneous tracking and jamming} (CSTJ) framework in which a team of autonomous UAVs cooperate in order to continuously track and jam a rogue drone in the air, thus forcing it to enter failsafe mode \cite{Revill2016} and auto-land or return to its base. More specifically, we focus on a realistic scenario in which the rogue drone moves in 3D space with uncertain and noisy dynamics. Each UAV is equipped with a 3D range-finding sensor \cite{Park2001} which returns noisy target measurements. In addition, we assume that the UAVs exhibit a limited sensing range and that they can detect the presence of the rogue drone inside their sensing range with probability less that 1. Due to imperfections of the 3D range-finding sensor we assume that, in addition to the target measurements the UAVs receive multiple false-alarm measurements at each time-step. Finally, we assume that the UAVs are equipped with a directional antenna which they use to transmit power to the rogue drone at discrete power-levels. 

\begin{figure}
	\centering
	\includegraphics[width=\columnwidth]{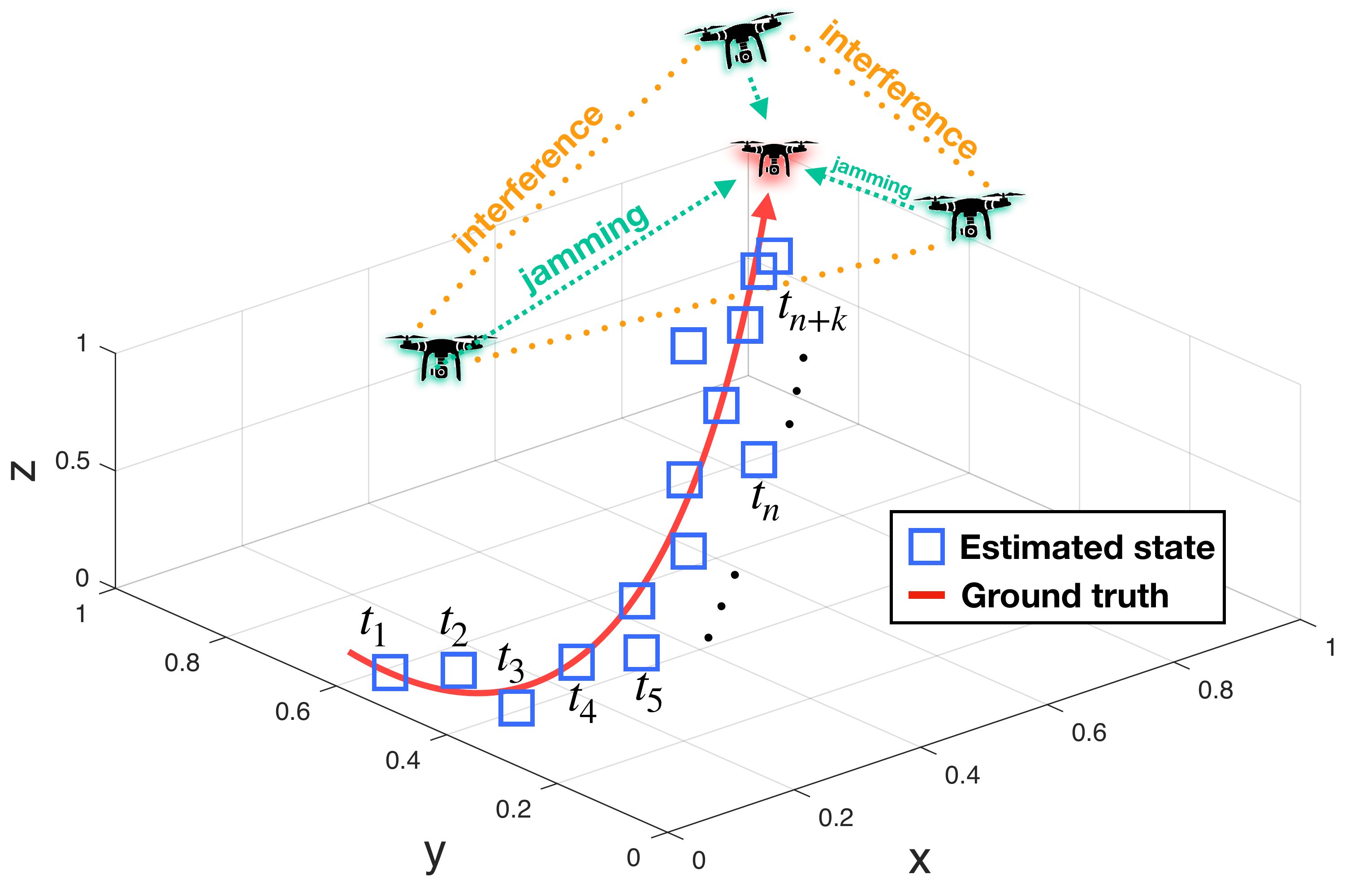}
	\caption{The figure illustrates the problem of cooperative simultaneous tracking and jamming tackled in this work.}	
	\label{fig:problem}
	\vspace{-0mm}
\end{figure}

In the proposed framework the UAVs cooperate in order to:  a) accurately track the rogue drone and b) choose the optimal transmit power levels from their on-board antennas which maximize the received power at the rogue drone and thus jamming its communications and sensing circuitry. At the same time the cooperative UAVs make sure that the power interference between them is kept below a specified critical interference threshold which in this work is assumed to be fixed and known. The main contributions of this work are the following:
\begin{itemize}
    \item We formulate the problem of cooperative simultaneous tracking and jamming of a rogue drone in challenging conditions e.g., uncertain dynamics, noisy measurements, false alarms, uncertain detection and with UAVs that exhibit limited sensing range. 
    \item We propose a novel decentralized estimation, decision and control framework that allows a team of cooperative UAVs to reposition themselves in 3D space at each time-step in order to accurately track-and-jam a rogue drone in the air, while at the same time avoiding the critical interference between them and thus remaining operational.
    \item We demonstrate the performance of the proposed approach through extensive simulation experiments.
\end{itemize}

\noindent The rest of the paper is organized as follows. Section \ref{sec:related_work} provides a brief overview of the existing literature on this topic by single and multiple agents. Section \ref{sec:system_model} develops the system model and Section \ref{sec:system_overview} formulates the problem and illustrates the proposed system architecture. Section \ref{sec:proposed_approach} discuses the details of the proposed approach and Section \ref{sec:Evaluation} conducts an extensive performance evaluation. Finally, Section \ref{sec:Conclusion} concludes the paper and discusses future  directions.

\section{Related Work} \label{sec:related_work}
A recent report by the U.S. Homeland Security Committee regarding the security threat posed by UAVs and possible countermeasures states that the interceptor UAV technology is currently immature and that interceptor UAVs equipped with reliable and accurate detection, tracking and defense capabilities are of essence \cite{Humphreys2015}. 
Initial works \cite{Guvenc2018,Ganti2016} on this problem investigate the detection and tracking of small UAVs using various methodologies including thermal, RF and audio/video signals. The work in \cite{Shi2018} provides a comprehensive overview of the technologies utilized for drone surveillance and discusses the state-of-the-art anti-drone systems. The works in \cite{Papaioannou2019_3,Hoffmann2006,Papaioannou2019_1,Papaioannou2019_2} investigate the problem of target tracking with single and multiple autonomous UAVs but without considering jamming capabilities. The work in \cite{Perkins2015} develops a UAV based solution for localizing a GPS jammer whereas the authors in \cite{Multerer2017} propose a low-cost ground jamming system that uses a 3D MIMO radar and a directional antenna to counteract the operation of small UAVs.
A more relevant work to the proposed system is described in \cite{Dressel2019} where a consumer UAV is outfitted with antennas and commodity radios in order to autonomously localize another drone using its telemetry radio signature. Moreover, the work in \cite{Brust2017}, develops an approach where a swarm of defense UAVs forms a cluster around a single rogue UAV in order to restrict its movement and escort it outside of the restricted airspace. The authors assume however, that a high-quality UAV monitoring system is in place which is able to accurately detect and track the rogue UAV. Finally, in \cite{Jang2017} the authors investigate the problem of cooperative control and task allocation in mission-planning and demonstrate their approach in radar jamming. 

Complementary to the aforementioned techniques, the proposed system investigates the problem of simultaneous tracking and jamming of a rogue drone with a team of cooperative autonomous UAVs. Compared to the existing techniques the proposed framework proposes a novel estimation, decision and control framework for the combined problem of tracking-and-jamming by multiple UAVs while considering the induced interference between them. This problem formulation, to the best of our knowledge has not been investigated before. 

%Extensive research has been pursued on target tracking \cite{Papaioannou2015,Papaioannou2017} and state estimation techniques \cite{Chen2003,Mahler2014book}, and comprehensive surveys can be found in \cite{Pulford2005,VoBa2015}. However, with the emergence of consumer drones the problem of target tracking has resurfaced and a variety of solutions for drone tracking have been proposed from academic and industrial research labs.
\section{System Model} \label{sec:system_model}

\subsection{Rogue Drone Dynamics}\label{ssec:target_dynamics}
Let us assume that the rogue drone evolves in 3D space with dynamics that can be expressed by the following discrete-time dynamical model:
\begin{equation} \label{eq:target_dynamics}
    x_t = \Phi x_{t-1} + \Gamma \nu_{t}
\end{equation}
where $x_t = [\text{x},\text{y},\text{z},\dot{\text{x}},\dot{\text{y}},\dot{\text{z}}]_t^\top \in \mathcal{X}$ denotes the drone's state at time $t$ which consists of the position and velocity components in 3D cartesian coordinates and $\nu_{t} \sim \mathcal{N}(0,\Sigma_\text{v})$ denotes the perturbing acceleration noise which is drawn from a zero mean multivariate normal distribution with covariance matrix $\Sigma_\text{v}$. The matrices $\Phi$ and $\Gamma$ are defined as:
\begin{equation}
\Phi = 
\begin{bmatrix}
    \text{I}_3 & \Delta T \cdot \text{I}_3\\
    \text{0}_3 & \text{I}_3
   \end{bmatrix},
\Gamma = 
\begin{bmatrix}
    0.5\Delta T^2 \cdot \text{I}_3\\
     \Delta T \cdot \text{I}_3
   \end{bmatrix}
\end{equation}

\noindent where $\Delta T$ is the sampling period, $\text{I}_3$ is the identity matrix of dimension $3 \times 3$ and $\text{0}_3$ is the zero matrix of dimension $3 \times 3$. In this work we assume that the drone dynamics obey the Markov property i.e., the state of the drone at the next time step depends only upon the state of the previous time step as shown in Eqn. (\ref{eq:target_dynamics}).

\subsection{UAV Dynamics} \label{ssec:AgentDynamics}
Suppose that we have in our disposal a set of controllable UAV agents $S = \{1,2,...,|S|\}$, where $|S|$ denotes the cardinality of the set, i.e., the number of available UAVs. Each UAV $j \in S$ is subject to the following discrete time dynamics:
\begin{equation} \label{eq:controlVectors}
s^j_{t} = s^j_{t-1} + \begin{bmatrix}
						\Delta_R[l_1] \sin(l_2 \Delta_\phi) \cos(l_3 \Delta_\theta)\\
						\Delta_R[l_1] \sin(l_2 \Delta_\phi) \sin(l_3 \Delta_\theta)\\
						\Delta_R[l_1] \cos(l_2 \Delta_\phi)
					\end{bmatrix},  
					\begin{array}{l} 
					    l_1 = 1,...,|\Delta_R|\\
					    l_2 = 0,...,N_\phi\\ 
						l_3 = 1,...,N_\theta
				    \end{array} 
\end{equation}
where  $s^j_{t-1} = [s^j_\text{x},s^j_\text{y},s^j_\text{z}]^\top_{t-1} \in \mathbb{R}^3$ denotes the state (i.e., position) of UAV $j$ (i.e., $(x, y, z)$ coordinates) at time $t-1$, $\Delta_R$ is a vector of possible radial step sizes ($\Delta_R[l_1]$ returns the value at index $l_1$), $\Delta_\phi=\pi/N_\phi$, $\Delta_\theta=2\pi/N_\theta$ and the parameters $(|\Delta_R|,N_\phi,N_\theta)$ determine the number of possible control actions. We denote the set of all admissible control actions of UAV $j$ at time $t$ as $\mathbb{U}^j_{t}=\{s^{j,1}_{t},s^{j,2}_{t},...,s^{j,|\mathbb{U}_{t}|}_{t}  \}$ as computed by Eqn. (\ref{eq:controlVectors}). 

%An illustrative example is shown in Fig. \ref{fig:controls}. We should point out that in this work we assume that the UAVs are capable of moving faster than the rogue drone and thus they are able to intercept it. 

\subsection{UAV Sensing Model} \label{ssec:agent_sensing}
The UAVs exhibit a limited sensing range for detecting a target, which is modeled by the function $p_D(x_t,s_t)$ that gives the probability that a target with state $x_t=[\text{x},\text{y},\text{z},\dot{\text{x}},\dot{\text{y}},\dot{\text{z}}]_t^\top$ at time $t$ is detected by the UAV with state $s_t=[s_\text{x},s_\text{y},s_\text{z}]^\top_t$. More specifically a target with state $x_t$ and position coordinates $Hx_t$ (where $H$ is a matrix that extracts the $(x, y, z)$ coordinates of a target from its state vector) is detected by a UAV with state $s_t$ with probability which is given by:
\begin{equation}\label{eq:sensing_model}
 p_D(x_t,s_t) = 
  \begin{cases} 
   p^{\max}_D & \text{if } d_t < R_0 \\
   \text{max}\{0, p^{\max}_D - \eta(d_t-R_0) \} & \text{if } d_t \ge  R_0
  \end{cases}
\end{equation}
where $d_t=\norm{H x_t-s_t}_2$ denotes the Euclidean distance between the UAV and the target in 3D space, $p^{\max}_D$ is the detection probability for a target which resides within $R_0$ distance from the UAV's position and finally $\eta$ captures the reduced effectiveness of the UAV to detect a distant target. 

Additionally, each UAV is equipped with a 3D range finding sensor which provides noisy measurements $y_t=[\rho,\theta,\phi]^\top_t \in \mathcal{Y}$ (i.e., radial distance $\rho$, azimuth angle $\theta$ and inclination angle $\phi$) from the detected target $x_t$ according to the following measurement model:
\begin{equation}\label{eq:meas_model}
    y_t = h(x_t,s_t) + w_t
\end{equation}
where $h(x_t,s_t)$ is given by:
\begin{equation}
    \left[ \norm{H x_t-s_t}_2, ~ \tan^{-1}(\frac{\delta_y}{\delta_x}), ~ \tan^{-1}(\frac{\sqrt{\delta_x^2+\delta_y^2}}{\delta_z}) \right]^\top
\end{equation} 
where $\delta_y = \text{y}-s_y$, $\delta_x = \text{x}-s_x$, $\delta_z = \text{z}-s_z$ and $w_t \sim \mathcal{N}(0,\Sigma_w)$ is zero mean Gaussian measurement noise with covariance matrix $\Sigma_w=\text{diag}[\sigma^2_\rho, ~\sigma^2_\theta, ~\sigma^2_\phi]$ and $\sigma_\rho$ is range dependent and given by $\sigma_\rho = \sigma_{\rho_0} + \beta_\rho \norm{H x_t-s_t}_2$.

Finally, due to the imperfection of the 3D range-finding sensor, at time $t$, a UAV agent receives (in addition to the target measurement) with an average Poisson rate of $\lambda_c$ a number of false alarm measurements $c^1_t,\ldots,c^n_t \in \mathcal{Y}, ~\mathbb{E}(n) = \lambda_c$ which are uniformly distributed inside the measurement space according to the density function $p_c(c_t)$. To summarize, at each time-step each UAV receives a collection of measurements: 

\begin{equation}\label{eq:Upsilon}
    \Upsilon_t=\bigcup \Big\{ a\subset\{ \emptyset,y_t\}, b \subseteq \{c^1_t,\ldots,c^n_t\} \Big\}
\end{equation}

\subsection{UAV Wireless Propagation Model} \label{ssec:agent_antena}
\begin{figure}
	\centering
	\includegraphics[scale=0.6]{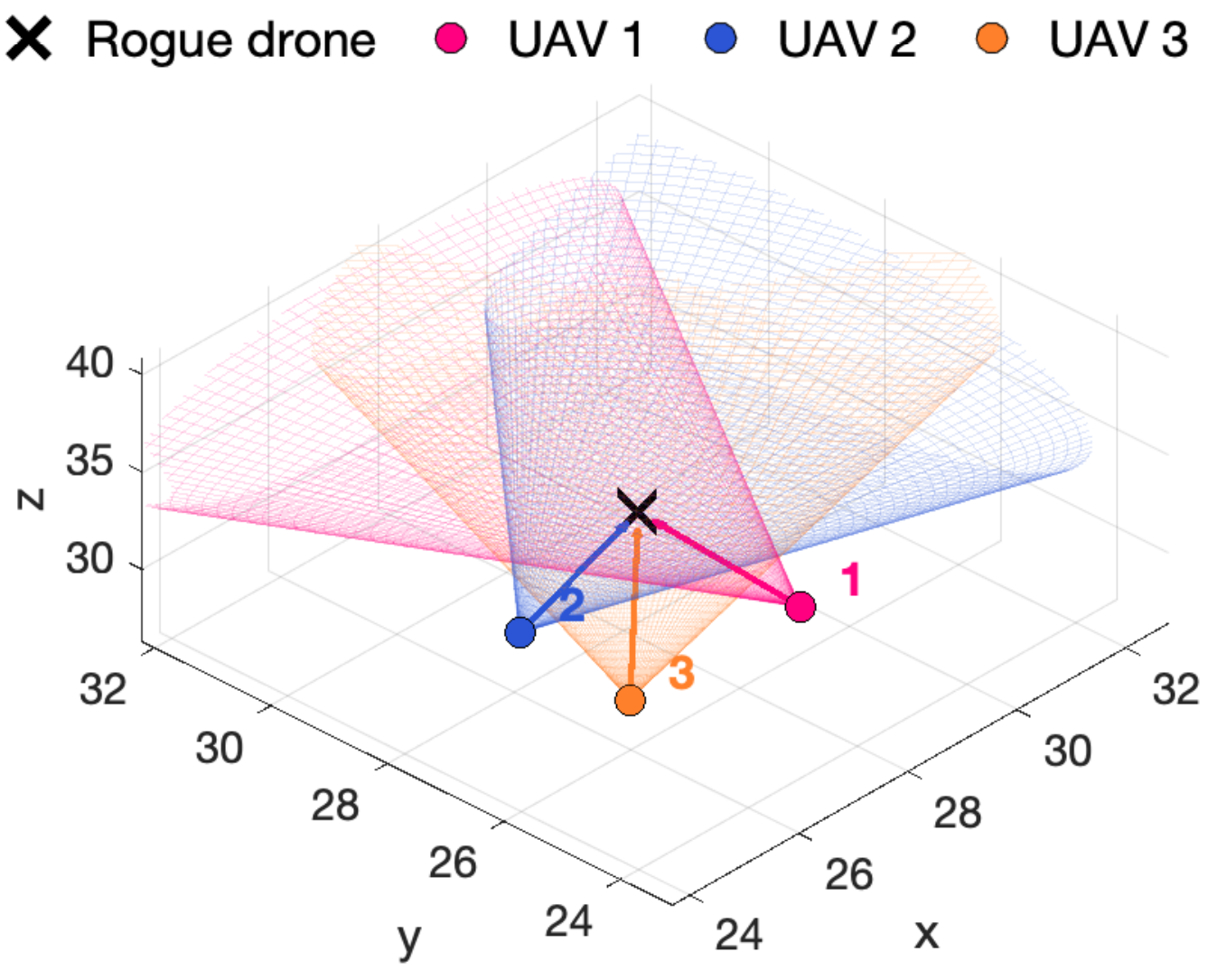}
	\caption{The figure illustrates the agent antenna model used in this work and described in Sec. \ref{ssec:agent_antena}. In the illustrative example above 3 UAV agents aiming their antennas (with $\theta_a = 60\deg$ and $h = 15$m) at the position of the target (i.e., rogue drone) to be jammed.}
	\label{fig:antenna}
	\vspace{-5mm}
\end{figure}

Each UAV carries a directional antenna which is used as the main mechanism for jamming the rogue drone. The antenna profile in cartesian 3D coordinates is described by a circular right angle cone which is given by: $[ x=u\cos(v), ~ y=u\sin(v), ~ z=\frac{h_a}{r} u ]$ for $ u \in [0, r], ~ v \in [0, 2\pi]$ where $h_a$ characterizes the effective antenna range, $r=\tan(\frac{\theta_a}{2})h_a$ is the circular base radius and $\theta_a$ is the opening angle of the cone which determines the antenna's conic lobe $\mathcal{A}$. At any point in time $t$, the antenna is auto-rotated to point in the direction of the vector $\vec{e}_t = \tilde{x}_t - s_t$, where $\tilde{x}_t$ and $s_t$ are the predicted position of the rogue drone and the state of the UAV, respectively. This is illustrated in Fig. (\ref{fig:antenna}).
The UAV emitted transmit power varies over a discrete set of power levels, $P^j_{t,\omega}, ~\omega\in\{1,\dots,\Omega\}, j \in S$ while the received power at the rogue drone $x_t$ from a UAV with state $s_t$ is governed by the following path-loss model:
\begin{equation}\label{eq:pathLoss}
\Lambda(s_t,x_t)=EF+10n_e\text{log}_{10}\left ( \norm{s_t-x_t}_2 \right)+AF
\end{equation}

\noindent where the term $EF$ models the near field loss effects, $n_e$ is the path-loss exponent and $AF$ models the attenuation effects \cite{Multerer2017}. We should point out here that due to the directional antenna of the UAVs, Eqn. (\ref{eq:pathLoss}) is valid only when $x_t$ resides inside the conic lobe $\mathcal{A}$ of the antenna of the UAV. When the rogue drone $x_t$ resides outside of $\mathcal{A}$, it receives no power from the UAV.

\section{Problem Statement and System Overview}\label{sec:system_overview}
In this paper we are interested in making optimized decisions on the UAV mobility controls ($u_t^{j} \in \mathbb{U}_t^j, \forall j \in S$) and UAV transmit power levels ($P^j_{t,\omega}, \forall j \in S$) of the cooperative UAVs in order to counteract the operation of a rogue drone. The problem of cooperative simultaneous tracking and jamming (CSTJ) tackled in this work can now be stated as follows: \textit { At each time step $t$ we would like to find the optimal joint UAV mobility control actions $u_t^{j} \in \mathbb{U}_t^j, \forall j \in S$ and at the same time choose the transmit power level $P^j_{t,\omega}, \forall j \in S$ for each UAV agent so that a) the UAVs maintain tracking of the rogue drone, b) the received power at the rogue drone is maximized and c) the in-between UAV interference is limited below a specified critical threshold.}

 A closer look at this problem however easily reveals its combinatorial complexity. Suppose we have in our disposal $|S|$ autonomous UAVs, each of which exhibits $|\mathbb{U}_t|$ mobility control actions and transmits in $\Omega$ different power levels. Then the number of joint mobility and power control actions that need to be evaluated at each time-step is given by $(|\mathbb{U}_t| \Omega)^{|S|}$, which quickly becomes computationally prohibitive to compute.
 
%For instance, even for relatively small number of UAVs (e.g. $|S|=4$), mobility controls (e.g. $|\mathbb{U}_t|=10$) and power levels (e.g. $\Omega=5$) we need to evaluate $625 \cdot 10^4$ combinations in each time-step to achieve our CSTJ objective. 

\begin{figure}
	\centering
	\includegraphics[width=\columnwidth]{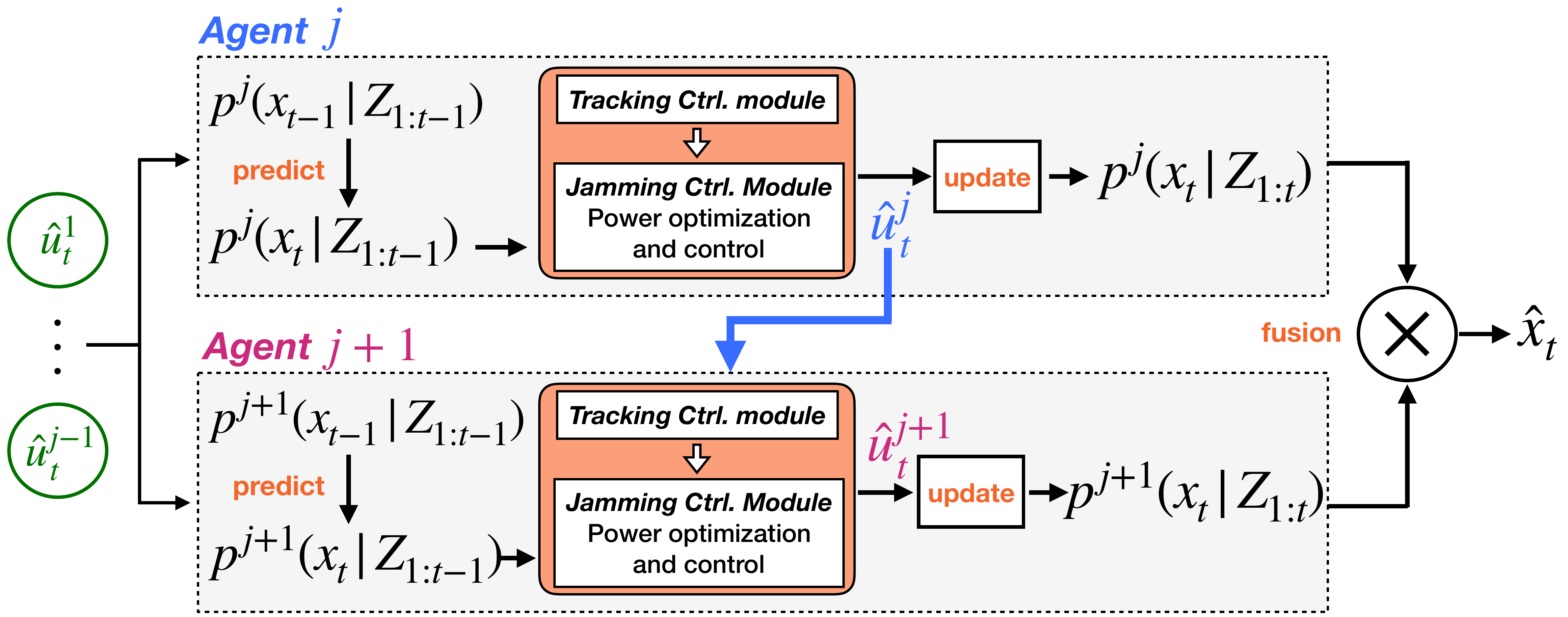}
	\caption{The figure illustrates the proposed cooperative simultaneous tracking and jamming (CSTJ) system architecture.}	
	\label{fig:sys_arch}
	\vspace{-3mm}
\end{figure}

For this reason, instead of tackling the above joint problem we propose an alternative suboptimal decentralized approach, as depicted in Fig. \ref{fig:sys_arch}, which allows us to solve instances of the problem in real-time. In essence we decouple the joint problem and we propose a cascaded control architecture which is composed of a \textit{Tracking control module} which gives input to a \textit{Jamming control module}. This cascaded controller is used by each UAV to determine its own optimal mobility and power control actions at each time step. Then we process the UAVs in a sequential fashion i.e., as it is shown in Fig. \ref{fig:sys_arch}, the UAV $j+1$ will take into consideration the actions taken by the already decided UAVs $1,\ldots,j$ when optimizing its CSTJ objective. 

Each UAV uses stochastic filtering to compute and propagate in time the probability distribution of the state of the rogue drone. Briefly, in stochastic filtering \cite{Simon2006,Chen2003} we are interested in the posterior filtering density $p(x_t|Z_{1:t})$ of some hidden state $x_t \in \mathcal{X}$ at time $t$ given all measurements up to time $t$ i.e., $Z_{1:t}=z_1,...,z_t$, with $z_t \in \mathcal{Z}$. Assuming an initial density on the state $p(x_0)$, the posterior density at time $t$ can be computed using the Bayes recursion as:
\begin{align} 
p(x_t|Z_{1:t-1}) &= \int  p(x_t|x_{t-1})  p(x_{t-1}|Z_{1:t-1}) d x_{t-1} \label{eq:predict}\\
p(x_t|Z_{1:t}) &= \frac{p(z_t|x_t)  p(x_t|Z_{1:t-1})}{\int p(z_t|x_t) p(x_t|Z_{1:t-1}) dx_t} \label{eq:update}
\end{align}
where Eqn. (\ref{eq:predict}) and (\ref{eq:update}) are referred to as the prediction and update steps respectively and the functions $p(x_t|x_{t-1})$ and $p(z_t|x_t)$ are the known transitional density and measurement likelihood function respectively. At each time step the hidden state $x_t$ is usually extracted from the posterior distribution using the expected a posteriori (EAP) or the maximum a posteriori (MAP) estimators. 
Finally, at each time-step the UAVs fuse their local estimation results to produce the final estimate for the state of the rogue drone.

\section{Cooperative Simultaneous Tracking and Jamming} \label{sec:proposed_approach}

\subsection{Tracking Control Module} \label{ssec:tracking_control}
In this subsection we describe the operation of the \textit{Tracking Control Module}, as depicted in Fig. \ref{fig:sys_arch}, which is responsible for actively controlling the movement of a UAV in order to maintain tracking of the rogue drone. More specifically, we seek to find the optimal mobility control action $u^j_t \in \mathbb{U}^j_t$ that must be taken at time-step $t$ by each UAV $j \in S$ so that the state of the rogue drone is estimated as accurately as possible. 

First, observe from the UAV sensing and measurement models (i.e., Eqn. (\ref{eq:sensing_model}) and Eqn. (\ref{eq:meas_model}) respectively) that the control action $u^j_t \in \mathbb{U}^j_t$ taken by UAV $j$ at time-step $t$ affects the probability of detecting the rogue drone and also the quality of the received measurements $y^j_t$. The state estimation of the rogue drone  depends on the received measurements which in turn depend on the applied control actions. 
For this reason we consider the UAV action which maximizes the probability of target detection (and thus the reception of target measurements) as our control strategy for maintaining tracking. That said, we denote the tracking control objective for agent $j$ as $\xi^j(x^j_t,u^j_t)$, and so the problem to solve becomes:
\begin{equation} \label{eq:track_control}
     \hat{u}_{t}^{j} = \underset{u^j_{t} \in \mathbb{U}^j_{t}}{\arg\max}~ \xi^j(x_t,u^j_t)=\underset{u^j_{t} \in \mathbb{U}^j_{t}}{\arg\max}~ p^j_D(x_t,u^j_t)
\end{equation}
where $x_t$ is the state the rogue drone. However, since $x_t$ is not available until the control action $\hat{u}^j_t$ is applied, we approximate it as $x^j_t \sim \tilde{x}^j_t$:
\begin{align}\label{eq:state_prediction}
    \tilde{x}^j_t &= \int x_t p^j(x_t|\Upsilon_{1:t-1}) d x_t
%    &= \int \int  x_t p(x_t|x_{t-1})  p^j(x_{t-1}|\Upsilon_{1:t-1}) d x_{t-1} d x_t \notag 
\end{align}

%\begin{align}\label{eq:state_prediction}
%    x_t \approx \tilde{x}_t &= {\arg\max}~ p^j(x_t|Y_{1:t-1})\\
%    &= {\arg\max} ~\int  p(x_t|x_{t-1})  p^j(x_{t-1}|Y_{1:t-1}) d x_{t-1} \notag 
%\end{align}
\noindent In essence each agent computes the predictive density $p^j(x_t|\Upsilon_{1:t-1})$ (i.e., Eqn. (\ref{eq:predict})) and extracts the expected predicted target state for time $t$. Note that in Eqn. (\ref{eq:state_prediction}) we have used $\Upsilon_t$ to denote the UAV measurements as described by Eqn. (\ref{eq:Upsilon}). In Sec. \ref{ssec:rfs_estimation} we discuss in more detail how we use $\Upsilon_t$ to estimate the state of the rogue drone.

\subsection{Jamming Control Module}
The solution of the problem presented in Eqn. (\ref{eq:track_control}) provides the optimal mobility control actions but only with respect to target tracking. These mobility control actions however, are not necessarily optimal for the objective of jamming. More specifically, in order to jam the rogue drone, the agents need to transmit a specific amount of power towards the target while at the same time ensuring that the power interference between them is kept below a predefined critical threshold $\Delta$. This is necessary so that the cooperative agents do not interfere with each other and remain operational at all times. In order to achieve the joint objective of tracking and jamming of the target, we utilize a cascaded control architecture in which we first find the optimal mobility control actions which result in a satisfactory tracking accuracy and then we perform a second optimization step where we refine further these mobility actions to achieve jamming control. To do that, instead of finding the mobility control actions which maximize Eqn. (\ref{eq:track_control}), we first compute for each agent $j$ the set $\tilde{\mathbb{U}}^j_t=\{\tilde{u}^{j1}_t,\tilde{u}^{j2}_t,...\}$ of all mobility control actions, which satisfy:
%\begin{equation}\label{eq:tracking_accuracy}
%    \tilde{\mathbb{U}}^j_t =  \underset{u^j_t \in \mathbb{U}^j_t}{\text{find}} ~\xi^j(\tilde{x}^j_t, u^j_{t}) > \vartheta 
%\end{equation}
\begin{equation}\label{eq:tracking_accuracy}
    \tilde{\mathbb{U}}^j_t =  \{u^j_{t} \in\ \mathbb{U}^j_t ~|~ \xi^j(\tilde{x}^j_t, u^j_{t}) > \vartheta \}
\end{equation}

\noindent where $\vartheta \in [0,1]$ is the desired threshold for the tracking objective. In the second step, we select the optimal mobility control actions and the levels of transmit power for each of the cooperative UAVs which maximize the received jamming power at the rogue drone.

To do that, each UAV decides in a sequential fashion its mobility and power control actions that maximize the received power at the rogue drone while satisfying the interference constraints on the power received from all previously decided UAVs. As discussed in Sec. \ref{ssec:agent_antena}, we consider a predefined discrete set of power levels $\Omega$ form which a UAV can chose its transmit power. Let $D$ be the set of all UAVs that have made a decision on their actions and $\bar{D}$ the set of undecided UAVs. Then the optimal decision and control problem for UAV $j$ can be formulated according to (P1) below:
\begin{align}
\text{(P1)}~\max & \sum_{\omega=1}^{\Omega}\sum_{k=1}^{|\mathbb{\tilde{U}}_t^j|} b_{\omega}^k \mathds{1}_{\mathcal{A}^j}(\tilde{x}^j_t) P_{t,\omega}^j\Lambda(\tilde{u}^{jk}_t,\tilde{x}^j_t) \label{eq:objective}\\
\mathrm{s.t.} &\>\>\sum_{\omega=1}^{\Omega}\sum_{i\in D} b_{\omega}^k \mathds{1}_{\mathcal{A}^i}(\tilde{u}^{jk}_t) \hat{P}_t^i\Lambda(\hat{u}^i_t,\tilde{u}^{jk}_t)<\Delta^j \label{eq:ownInterference}\\
&\sum_{\omega=1}^{\Omega}\sum_{k=1}^{|\mathbb{\tilde{U}}_t^j|} b_{\omega}^k \mathds{1}_{\mathcal{A}^j}(\hat{u}^{i}_t) P_{t,\omega}^j\Lambda(\tilde{u}^{jk}_t,\hat{u}^i_t) ~~+ \label{eq:otherInterference}\\
&\>\> \sum_{l\in D,l\neq i}  \mathds{1}_{\mathcal{A}^l}(\hat{u}^{i}_t)\hat{P}_t^l\Lambda(\hat{u}^l_t,\hat{u}^i_t)<\Delta^i,~\forall i\in D \notag \\
& \>\> b_\omega^k\in\{0,1\}, \>\> \sum_{\omega=1}^{\Omega}\sum_{k=1}^{|\mathbb{\tilde{U}}_t^j|}  b_{\omega}^k = 1 \label{eq:chooseCombination}
\end{align}

%In $(P1)$ is a combinatorial problem where the best alternative mobility control action is selected and the transmit power chosen by each agent in order to maximize the target received power while ensuring that the interference level between cooperating agents is maintained below thershold $\Delta$. Constrained eq. \eqref{eq:ownInterference} ensures that the interference caused to agent $j$ when taking action $\tilde{u}^{kj}_t$ from those agents that have already made their decisions is below the threshold. On the other hand, eq. \eqref{eq:otherInterference} ensures that the interference caused by agent $j$ with $\tilde{u}^{kj}_t$ and $P_{\omega}^j$ in addition to the interference caused by previously decided agents is still within the acceptable interference limits for all $i\in D$. The selection of a single combination of control actions is ensured by eq.  \eqref{eq:chooseCombination}. 

\noindent In $\text{(P1)}$ the best mobility control action is selected and the transmit power is chosen for each UAV $j$ which maximizes the received power at the rogue drone while ensuring that the interference level between the cooperating UAVs is maintained below the critical threshold $\Delta^j$. The indicator function $\mathds{1}_{\mathcal{A}^j}(x)$ checks if $x$ is within the conic antenna lobe $\mathcal{A}^j$ of UAV $j$ and returns 1, otherwise returns 0.
The constraint in Eqn. (\ref{eq:ownInterference}) ensures that the interference caused to agent $j\in \bar{D}$ when taking action $\tilde{u}^{jk}_t \in \tilde{\mathbb{U}}^j_t$ with respect to all UAVs ($i \in D \subseteq S$) that have already made their mobility and power-level decisions (i.e., $\hat{u}^i_t$ and $\hat{P}_t^i$) is below the critical threshold. 
On the other hand, Eqn. (\ref{eq:otherInterference}) computes a) the interference caused by UAV $j \in \bar{D}$ to UAV $i \in D$ that has already made a decision when UAV $j$ applies mobility control $\tilde{u}^{jk}_t$ and power-level $P_{t,\omega}^j$ and b) the interference caused to UAV $i$ by all UAVs $l \ne i \in D$ that have made a decision and transmit in power-level $\hat{P}_t^l$. Then, this constraint ensures that the interference caused to UAV $i \in D$ from all the already decided UAVs $l \in D$ and the UAV $j \in \bar{D}$ is still within the acceptable interference limit $\Delta^i$. Finally, the selection of a single combination of control actions (i.e., mobility and power-level) is ensured by Eqn. (\ref{eq:chooseCombination}) using the binary variable $b_\omega^k$.
In essence, $\text{(P1)}$ will find $\omega \in [1,\ldots,\Omega]$ and $k \in [1,\dots,|\tilde{\mathbb{U}}_t|]$ for each agent $j \in S$ such that the applied power-level $\hat{P}_t^j$ and mobility control $\hat{u}^j_t$ maximize Eqn. (\ref{eq:objective}) while satisfying the interference constraints. The sequential optimization approach of (P1) along with the proposed cascaded control architecture allow for the evaluation of fewer mobility control actions at each time-step and reduces the exponential complexity of the joint problem to linear in the number of agents and total control actions.

%{\color{red} Notably, only a single combination $b_\omega^k,\forall \omega$ is required in each time instance, hence $\text{(P1)}$ is a simple assignment problem that can be solved efficiently even for large instances.}

 %Solving $(P1)$ provides the best alternative mobility control action, $\tilde{\mathbb{U}}^j_t$ and the best alternative power level $P_\omega^j$ by each agent to obtain the maximum received powered at the target while satisfying the interference levels between the cooperating agents. 

\subsection{Rogue Drone State Estimation} \label{ssec:rfs_estimation}

The objective here is to use stochastic filtering in order estimate the state of the rogue drone given its uncertain dynamics, noisy measurements and  false-alarms as described in Sec. \ref{sec:system_model}. In order to do that we need to compute the expressions for the transitional density and measurement likelihood function  as discussed in Sec. \ref{sec:system_overview}.
From our modeling assumptions we already know that $p(x_t|x_{t-1}) = \mathcal{N}(x_t;\Phi x_{t-1},\Gamma\Sigma_\text{v}\Gamma^\top)$ is a multivariate Gaussian transitional density according to the target dynamics of Eqn. (\ref{eq:target_dynamics}).  What is left is to compute the expression for the likelihood function $p(z_t|x_t)$ (see Eqn. (\ref{eq:update})) which in our case becomes $p(\Upsilon_t|x_t,s_t)$. This can be computed as:
\begin{align} \label{eq:rfs_like}
    p(\Upsilon_t|x_t,s_t) &= \left[1-p_D(x_t,s_t)\right]e^{-\lambda_c} \prod_{\ell \in \Upsilon_t} \lambda_c p_c(\ell) ~+  \\
   & e^{-\lambda_c}  p_D(x_t,s_t) \underset{\ell \in \Upsilon_t}{\sum} g_y(\ell|x_t,s_t) \underset{ \begin{subarray}{l}
  \varepsilon \in \Upsilon_t \\
  \varepsilon \ne \ell
  \end{subarray}}{\prod} \lambda_c p_c(\varepsilon) \notag
\end{align}
\noindent where the function $g_y(y_t|x_t,s_t) = \mathcal{N}(y_t;h(x_t,s_t),\Sigma_w)$ is due to Eqn. (\ref{eq:meas_model}). To obtain the above expression we first use the fact that if the UAV receives only false alarm measurements (i.e., no measurement from the rogue drone) then $\Upsilon_t=\{c^1_t,...,c^{|\Upsilon_t|}\}$ and $ p(\Upsilon_t|x_t,s_t)$ is given by:
\begin{equation}
    |\Upsilon_t|! \left[1-p_D(x_t,s_t)\right] \text{Pois}(|\Upsilon_t|) \prod_{\ell \in \Upsilon_t} p_c(\ell)
\end{equation}

\noindent where the function $\text{Pois}(|\Upsilon_t|) =\frac{\lambda_c^{|\Upsilon_t|}e^{-\lambda_c}}{|\Upsilon_t|!}$ is the Poisson distribution with rate parameter $\lambda_c$ and gives the probability of obtaining exactly $|\Upsilon_t|$ false alarm measurements at time $t$ and $\prod_{\ell \in \Upsilon_t} p_c(\ell)$ is the joint spatial density of the false alarms. The above expression is multiplied by $|\Upsilon_t|! [1-p_D(x_t,s_t)]$ to account for all possible permutations of the measurement sequence and for the fact that the target has not been detected, which explains the first line in Eqn. (\ref{eq:rfs_like}).

The second line of Eqn. (\ref{eq:rfs_like}) is due to the more general case in which the UAV receives measurements from the rogue drone $y_t$ in addition to false alarms e.g., $\Upsilon_t=\{y_t,c^1_t,c^2_t,\ldots\}$. In this case the joint likelihood accounts for the detection of the rogue drone and its measurement i.e., $p_D(x_t,s_t) g_y(y_t|x_t,s_t)$, multiplied by the joint density of the remaining $|\Upsilon_t|-1$ measurements which are considered false alarms (i.e., $(|\Upsilon_t|-1)! \cdot \text{Pois}(|\Upsilon_t|-1)~\prod_{\ell~\in~\Upsilon_t,~\ell~\ne~y_t}~p_c(l)$). Finally, we have to sum over all possible $|\Upsilon_t|$ ways in which we can select $y_t$ from $\Upsilon_t$.

\begin{figure*}
	\centering
	\includegraphics[width=\textwidth]{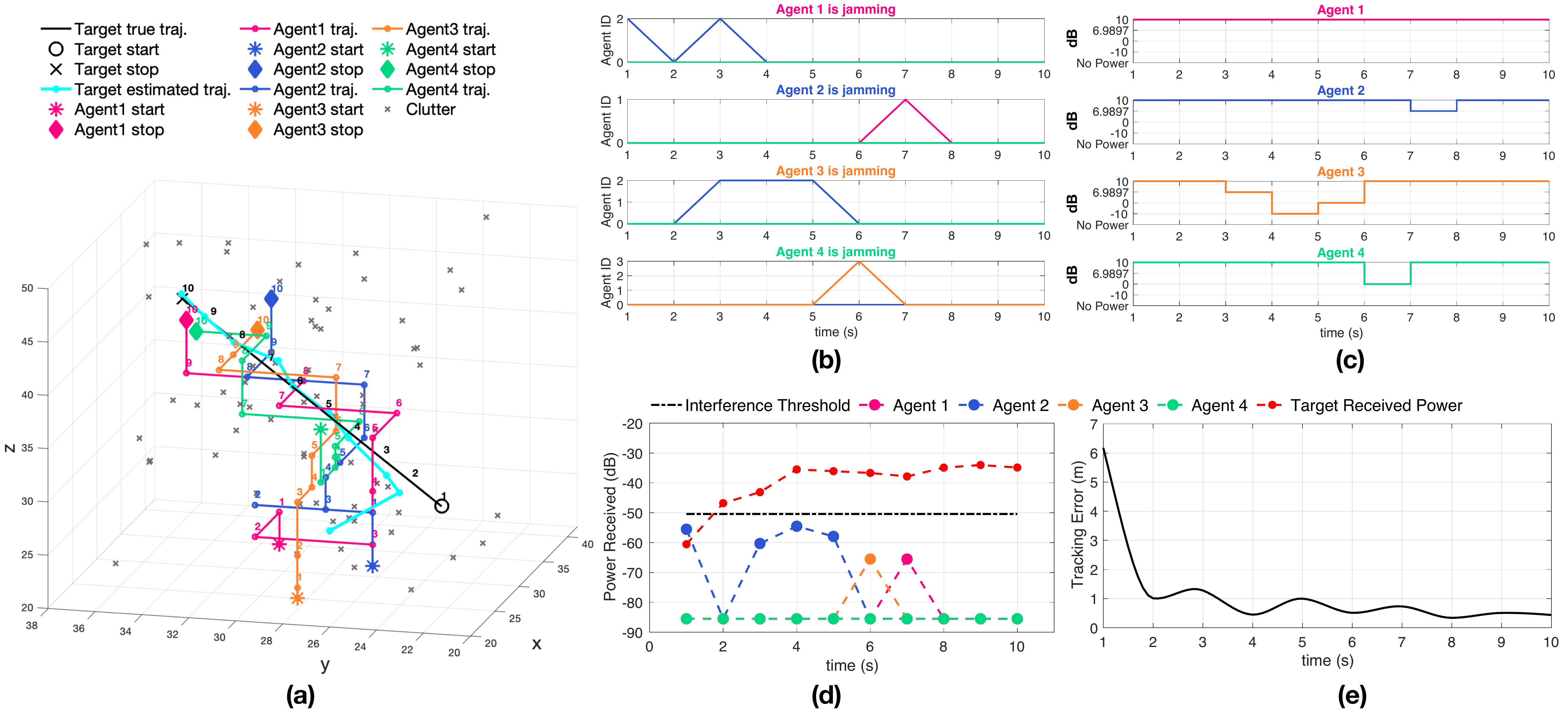}
	\caption{The figure shows the trajectories of 4 agents for the task of cooperative simultaneous tracking and jamming of a rogue drone.}	
	\label{fig:scenario}
	\vspace{-1mm}
\end{figure*}

To summarize each UAV $j$ propagates in time the filtering density $p^j(x_t|\Upsilon_{1:t})$ and computes the EAP state for the rogue drone as $\hat{x}^j_t = \int x_t p^j(x_t|\Upsilon_{1:t}) dx_t$ and its covariance matrix $K^j_{\hat{x}^j_t} = \mathbb{E}[(x-\hat{x}^j_t)(x-\hat{x}^j_t)^\top]$. Finally, the UAVs exchange their local estimates with each other and apply sequential covariance intersection \cite{Deng2012} to compute the final fused state $\hat{x}_t$ of the rogue drone. The resulting CSTJ algorithm is shown below in Alg. \ref{alg:Algo1}.

%This is useful for the operation of the proposed system since it allows the agents to have an accurate image of where the target is located. There are cases in where the agents happen to be away from the target (e.g. in order to reduce interference). The sensor fusion helps these agents to better maintain tracking of the target in such situations. The resulting CSATJ control algorithm is presented below.

\begin{algorithm}
\caption{Proposed Cooperative Simultaneous Tracking and Jamming approach}
\label{alg:Algo1}
\begin{algorithmic}[1]
\REQUIRE $p^j(x_{t-1}|\Upsilon_{1:t-1}), \mathbb{U}^j_t,P^j_{t,\omega} ~ \forall ~j,\omega, D \subseteq S$
\STATE Compute the predictive density $p^j(x_{t}|\Upsilon_{1:t-1})$ using Eq. (\ref{eq:predict}).
\STATE Compute the predicted target state at time $t$, $\tilde{x}^j_t$ using Eqn. (\ref{eq:state_prediction})
\STATE Obtain the set of optimized tracking control actions $\tilde{\mathbb{U}}^j_t$ by solving Eqn. (\ref{eq:tracking_accuracy})
\STATE Solve P1 to obtain the optimal mobility control action $\hat{u}^j_t$ and power-level $\hat{P}^j_t$.
\STATE Execute the optimal mobility and power control actions.
\STATE Receive the target measurement set $\Upsilon^j_t$.
\STATE Compute posterior density $p^j(x_{t}|\Upsilon_{1:t})$ by plugging in Eqn. (\ref{eq:rfs_like}) into Eqn. (\ref{eq:update})  
\STATE Estimate target state $\hat{x}^j_t$ and its covariance matrix $K^j_{\hat{x}^j_t}$.
\STATE Produce the final state estimate $\hat{x}_t$ by fusing together the information from line 8 using sequential covariance intersection \cite{Deng2012}.

\end{algorithmic}
\end{algorithm}

\section{Evaluation}\label{sec:Evaluation}

\subsection{Simulation Setup}

In order to evaluate the performance of the proposed approach we have conducted several numerical experiments. For these experiments we have used the following setup: The UAV agents and the rogue drone maneuver in a 3D space with dimensions $100\text{m} \times 100\text{m} \times 100\text{m}$. The target dynamics are according to Eqn. (\ref{eq:target_dynamics}) with $\Sigma_\text{v}=\text{diag}[2,2,2] \text{m}/\text{s}^2$ and $\Delta T = 1$s. The agent dynamics are according to Eqn. (\ref{eq:controlVectors}) with $\Delta_R = [1,  3, 5]$, $N_\phi = 2$ and $N_\theta= 4$. The agent sensing model is given by Eqn. (\ref{eq:sensing_model}) with $p_D^{\text{max}} = 0.99$, $\eta = 0.02 \text{m}^{-1}$ and $R_0=2$m. The agent's measurement model is given by Eqn. (\ref{eq:meas_model}) with parameters $\sigma_\theta = ~\sigma_\phi = \pi/50$rad, $\sigma_{\rho_0} = 2$m and $\beta_\rho = 0.05 \text{m}^{-1}$ and finally $\lambda_c = 15$. For the antenna model we have $h_a=100$m and $\theta_a=80$deg. Additionally, the UAV agents transmit in the following power levels $[\text{off}, -10, 0, 7, 10]$dB, the agent interference level $\Delta$ is set at $-50$dB, the path-loss exponent is $2.5$, the near field loss effects $EF$ are set to 32.4dB and the attenuation effects $AF$ are computed based on a $2$GHz carrier frequency as $AF = 20\text{log}_{10}(2) = 6.0206$dB. Finally, the tracking threshold $\vartheta$ is set to $\vartheta=0.8$ and the filtering approach described in Sec. \ref{ssec:rfs_estimation} is implemented as a SIR particle filter \cite{Gustafsson2010} in order to handle the non-linear sensing model described in Sec. \ref{ssec:agent_sensing}.

\subsection{Results}
Figure \ref{fig:scenario}a shows a simulated scenario over 10 time-steps in which 4 agents cooperate in order to simultaneously actively track and jam a single target. More specifically, agents 1, 2, 3 and 4 (shown in different colors) are spawned from $[29,30,23], [24,25,24], [23,28,21]$ and $[23,27,37]$ respectively. The target initial state is $[21,21,30,1.8,1.6,1.1]$ as indicated by the black circle and moves along the black line with final position marked with $\times$ as shown in the figure. The agents maneuver around the target in order to achieve their CSTJ objective. More specifically, the agents cooperatively choose their control actions in each time step which result in the optimal tracking and jamming performance while at the same time accommodating their in-between interference constraints. Interestingly, to achieve this the agents follow interleaving trajectories to avoid the interference between them. Moreover, it is shown that these optimized control decisions also allow for a satisfactory tracking performance. As time progresses the estimated target trajectory (i.e., cyan line) converges to the ground truth (i.e., black line). This is also evident by the tracking error (i.e., the Euclidian distance between the estimated and the ground truth target position) shown in Fig. \ref{fig:scenario}e. In more detail, the agents try to maximize the target received power while at the same time keeping the interference between them below the critical threshold $\Delta$. In order to meet the interference constraints the agents can either (a) optimize their mobility controls and move to a non-interfering position or (b) optimize the antenna power-level and possibly transmit at a lower power-level. At the same time the agents need to accurately track the target over time in order to be able to steer their antenna towards the direction of the target. Figure \ref{fig:scenario}b shows the agents being jammed (due to interference), Fig. \ref{fig:scenario}c shows at which power-levels the agents transmit power and finally Fig. \ref{fig:scenario}d shows the target received power and the amount of received power for each agent due to interference. As we can observe at time-step $t=1$, agent 1 is jamming agent 2, and transmits with full power at 10dB. Agent 2 however, manages to move to a position at $t=1$ in which the interference caused by agent 1 remains below the critical threshold as shown in Fig. \ref{fig:scenario}d. Moreover, at the beginning of this experiment (i.e., $t=1..3$) the target received power is relatively low compared to later time-steps (Fig. \ref{fig:scenario}d). This is because, at the beginning of the experiment the target estimated position deviates significantly from the true position (due to the high initial uncertainty on the target position) and thus the agents are having difficulty steering their antennas at the direction of the target. This however, changes as the agents improve their cooperative estimation over time. Moreover, it is shown that between time-steps $t=3..5$ agent 3 is jamming agent 2 and for this reason agent 3 adjusts its power-levels accordingly in order to keep the received power at agent 2 below the critical threshold as shown in Fig. \ref{fig:scenario}c and Fig. \ref{fig:scenario}d. What we have discussed so far briefly illustrates the operation of the proposed system. We should point out here that the parameters of the agent dynamical model i.e., Eqn. (\ref{eq:controlVectors}), used for this simulation have been chosen in order to produce the illustrated Manhattan-like trajectories in order to aid visual inspection and analysis.

\begin{figure}
	\centering
	\includegraphics[width=\columnwidth]{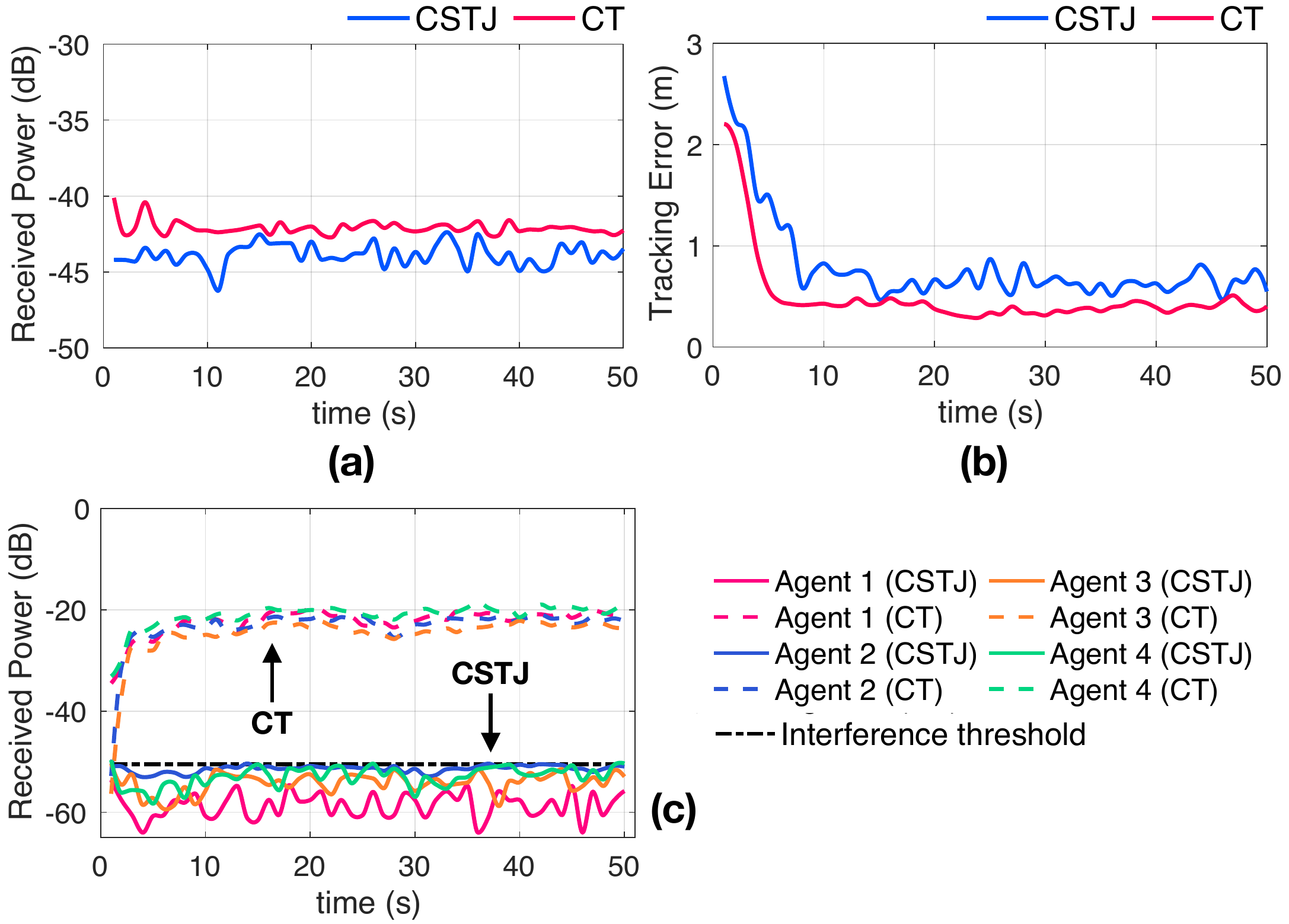}
	\caption{Performance comparison between CSTJ and CT.}	
	\label{fig:stats1}
	\vspace{-7mm}
\end{figure}

Next, we compare the performance of the proposed system (CSTJ) against a baseline method which performs cooperative tracking (CT) with jamming capabilities. In CT the UAVs are capable of jamming the rogue drone, however the interference between the UAVs is not taken into account. Additionally, in CT each agent maximizes its own tracking objective as discussed in Sec. \ref{ssec:tracking_control} and then their local results are fused using covariance intersection in order to compute the final target state. This experiment aims to investigate the pros and cons of the proposed system in terms of tracking performance, jamming performance and the ability of the system to maintain the UAVs below the specified interference threshold. Since, in CT the UAVs do not posses the capability of switching/deciding between the different transmit power levels, for this experiment we assume that in both methods (CSTJ and CT) the UAV antenna is always on and transmits with constant power at 7db. 
For this test we used the following procedure: First, we randomly initialize a target inside the surveillance area. Then we spawn randomly 4 agents inside a sphere with radius 5m centered at the target birth location and we run CSTJ and CT for 50 time-steps monitoring the target received power, the tracking error and the interference between the UAVs. The above procedure is repeated 50 times. Figure \ref{fig:stats1} shows the average values of (a) target received power, (b) tracking error and (c) UAV interference for the 50 trials. More specifically, Fig. \ref{fig:stats1}a shows that the target received power obtained with CT is on average 1.7dB higher in each time-step compared to what is obtained with CSTJ. Moreover, with CT the tracking error shown in Fig. \ref{fig:stats1}b is on average 0.3m lower compared to the error achieved with CSTJ. These results are quite reasonable since with the addition of the jamming objective in CSTJ, the UAVs do not always take the optimal control actions that result in the best possible tracking performance. In contrast, they consider the joint tracking and jamming objective which might result in control decisions that deviate for the optimal in terms of tracking accuracy. In addition since CT does not considers any interference constraints between the agents, it usually achieves higher target received power as the agents can get closer to the target which explains the results obtained. However, the advantage of the proposed CSTJ framework is depicted in Fig. \ref{fig:stats1}c. More specifically, the proposed system can maintain the interference between the UAVs below the critical threshold of -50dB while achieving reasonable tracking and jamming performance. On the other hand the CT does not considers these constraints and as a result the system becomes unstable and fails as the UAVs receive very large amounts of power. This is because in CT the UAVs can get extremely close to each other.

\begin{figure}
	\centering
	\includegraphics[width=\columnwidth]{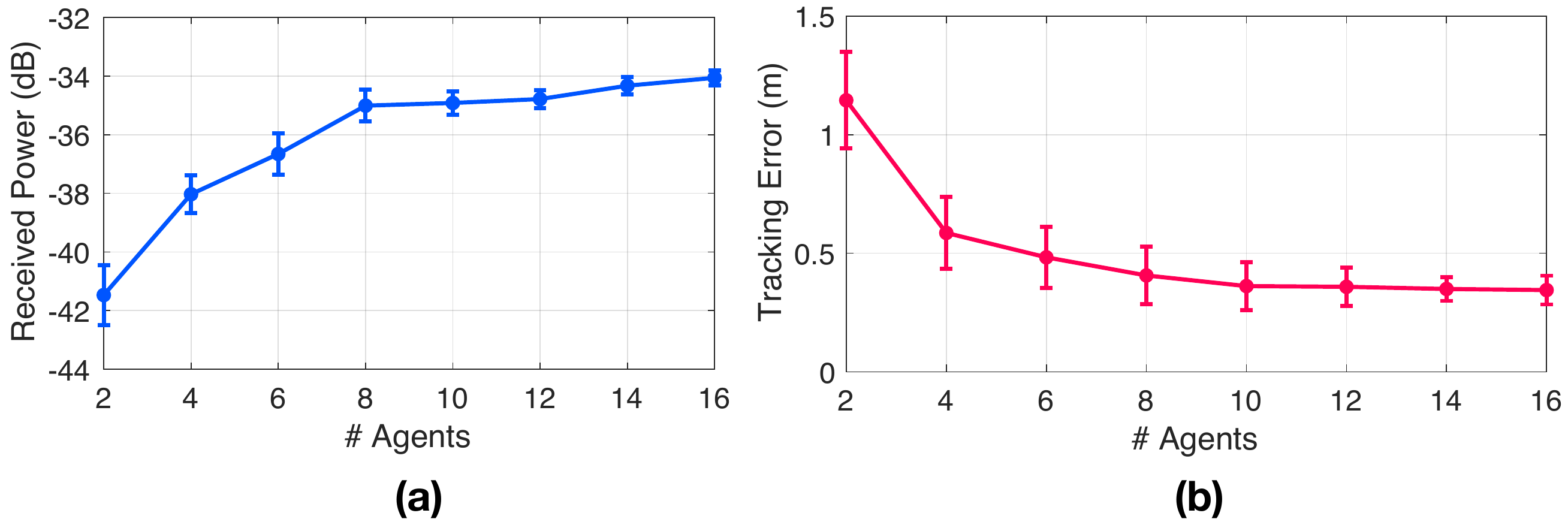}
	\caption{The figure shows the effect of the number of UAVs on the performance of the proposed system.}	
	\label{fig:stats2}
	\vspace{-7mm}
\end{figure}

Finally, we investigate how the number of agents affects the performance of the proposed approach. In order to do that we follow a similar setup with the one described in the previous paragraph. We randomly initialize a target inside the surveillance area and we vary the number of agents that are randomly spawned around the target (within a 5m radius). We let our system to run for 50 time-steps and we measure the average power received at the target and the average tracking error. Figure \ref{fig:stats2} shows the averaged results of this experiment over 50 Monte-Carlo trials. For this test we set the agent power-levels as $[\text{off}, 0, 7, 10]$dB and their mobility control parameters as $\Delta_R = [1, 3]$, $N_\phi = 2$ and $N_\theta= 4$.
 
 As we can observe from Fig. \ref{fig:stats2}a the average received power at the target increases significantly between 2 and 8 agents and then slows down. For more than 8 agents we can say that the target received power almost reaches a plateau. A similar pattern is also shown for the tracking error in Fig. \ref{fig:stats2}b. Interestingly, this behavior is due to the following reason: as the number of agent increases the number of available mobility control actions that the agents can find which result in good tracking performance and satisfy the interference constraints decreases. We have observed that as the number of agent increases, the agents either a) switch off their antenna and transmit no power since they cannot find a position to move into in which the interference constraints are met or b) move to a distant position (with respect to other UAVs) and transmit with minimum power. This limitation can be alleviated (with extra computational cost) by increasing the the number of degrees of freedom of the system i.e., increasing the number of mobility control actions, the power-levels and reducing the antenna opening angle $\theta_a$.

\section{Conclusion} \label{sec:Conclusion}
In this paper we have studied the problem of simultaneous tracking and jamming of a rogue drone with a team of cooperative UAV agents. We have presented a novel estimation, decision and control framework that allows the UAVs to reconfigure their positions at each time-step so that the tracking-and-jamming objective is achieved while taking into account the interference induced between them. Finally, we have demonstrated the effectiveness of the proposed technique through extensive simulation experiments. Future work, will focus on the real-world implementation of the proposed system, and its extension to multiple targets.

\section*{Acknowledgments}
This work has been supported by the European Union's H2020 research and innovation programme under grant agreement No 739551 (KIOS CoE) and from the Government of the Republic of Cyprus through the Directorate General for European Programmes, Coordination and Development.
\bibliographystyle{IEEEtran}
\bibliography{IEEEabrv,main}

\end{document}